# RECOGNIZING BANGLA GRAMMAR USING PREDICTIVE PARSER

## K. M. Azharul Hasan, Al-Mahmud, Amit Mondal, Amit Saha

Department of Computer Science and Engineering (CSE)
Khulna University of Engineering and Technology (KUET)
Khulna-9203, Bangladesh

azhasan@gmail.com, rocky_cse_05@yahoo.com, amit.kuet@gmail.com,
amit_kuet@yahoo.com

**Abstract-** *We describe a Context Free Grammar (CFG) for Bangla language and hence we propose a Bangla parser based on the grammar. Our approach is very much general to apply in Bangla Sentences and the method is well accepted for parsing a language of a grammar. The proposed parser is a predictive parser and we construct the parse table for recognizing Bangla grammar. Using the parse table we recognize syntactical mistakes of Bangla sentences when there is no entry for a terminal in the parse table. If a natural language can be successfully parsed then grammar checking from this language becomes possible. The proposed scheme is based on Top down parsing method and we have avoided the left recursion of the CFG using the idea of left factoring.*

**Index terms-** Context Free Grammar, Predictive Parser, Bangla Language processing, Parse Table, Top down and Bottom up Parser, Left Recursion.

## 1. INTRODUCTION

Parsing is the process of using grammar rules to determine whether a sentence is legal, and to obtain its syntactical structure. Tree structure provides two information viz. it divides the sentence into *constituents* (in English, these are called phrases) and it puts them into *categories* (Noun Phrase, Verb Phrase, etc).To process any natural language, parsing is the fundamental problem for both machines and humans. In general, the parsing problem includes the definition of an algorithm to map any input sentence to its associated syntactic tree structure[1]. A parser analyzes the sequence of symbols presented to it based on the grammar[2]. Natural language applications namely Information Extraction, Machine Translation, and Speech Recognition, need to have an accurate parser[3]. Parsing natural language text is more difficult than the computer languages such as compiler and word processor because the grammars for natural languages are complex, ambiguous and infinity number of vocabulary. For a syntax based grammar checking the sentence is completely parsed to check the correctness of it. If the syntactic parsing fails, the text is considered incorrect. On the other hand, for statistics based approach, Parts Of Speech (POS) tag sequences are prepared from an annotated corpus, and hence the frequency and the probability[4]. The text is considered correct if the POS-tagged text contains POS sequences with frequencies higher than some threshold[5].

Bangla is yet in degraded stage at least as far as work in the area of computational linguistics is concerned. Natural languages like English and even Hindi are rapidly progressing as far as work done in processing by computers is concerned. Unfortunately, Bangla lags more or less behind in some crucial areas of research like parts of speech tagging, text summarization and categorization, information retrieval and most importantly in the area of grammar checking. The grammar checking for a language has a wide variety of applications. Although Bangla is the fourth largest language of the world having over 200 million native speakers but still now Bangla language does not have a complete computerized grammar checker for a given Bangla sentence. In this paper, we proposed a context free grammar for the Bangla language and hence we proposed a predictive Bangla parser constructing a parse table. We have adopted the top down parsing scheme and avoided the problem of left recursion using left factoring for the proposed grammar. We implemented the Bangla dictionary in XML format using the corresponding word as tag name and it's POS as value. It is very much useful technique because it needs less time to/from data store or data retrieval from this data storage rather than other forms of data storage. It helps to search the dictionary very fast. Bangla grammar has huge amount of forms and rules .We believe the proposed grammar and parser can be applicable to any forms of Bangla sentences and can be used as grammar checker.

## 2. RELATED WORKS

A rule based Bangla parser has been proposed in [1] that handles semantics as well as POS identification from Bangla sentences and ease the task of handling semantic issues in machine translation. An open-source morphological analyzer for Bangla Language using finite state technology is described in [7]. They developed the monolingual dictionary called monodix and stored in XML file. [5] addresses a method to analyze syntactically Bangla sentence using context-sensitive grammar and interpret the input Bangla sentence to English using the NLP conversion unit. The system is based on analyzing an input sentence and converting into a structural representation. A parsing methodology for Bangla natural language sentences is proposed in [8] and shows how phrase structure rules can be implemented by top-down and bottom-up parsing approach to parse simple sentences of Bangla. A comprehensive approach for Bangla syntax analysis was developed [9] where a formal language is defined as a set of strings. Each string is a concatenation of terminal symbols. Some other approaches such as Lexical Functional Grammar (LFG) [4] and Context Sensitive Grammar (CSG) [10] –[12] have also been developed for parsing Bangla  sentences. Some developers devoloped Bangla parser using SQL to check the correctness of sentence; but its space complexity is inefficient. Besides it take more time for executing SQL command. As a result those Parser becomes slower.

A technique is implemented to perform structural analysis of Bangla sentences of different tenses using Context Free Grammar (CFG) rule [13]. A methodology for analysis the Bangla sentences in semantic manner is presented in [14] and [15] presents a technique for detecting the named entity based on classifier for Bangla documents. But these papers[13]-[15] does not deal with the detail grammar recognition for Bangla sentences.

```
<?xml version="1.0" encoding="ISO-8859-1"?>
<!-- Edited by XMLSpyⅡ-->
<WORD>
        <আমি>pronoun</আমি>
       <খাই>verb</খাই>
        <একটি>modifier</একটি>
         <এবং>conjunction</এবং>
       <আমরা>pronoun</আমরা>
       <না>neg</না>
</WORD>
```

Figure 1: Data Format of XML File

In this paper we store the Bangla word as tag and its corresponding constituents or POS are used as value. Hence to the POS or constituents of a word it will be speedy to get the result. More over we have applied the idea of left factoring to remove the left recursion and ambiguity of the grammar and hence according to our proposed grammar, the parser can detect the errors in a Bangla sentence.All the scheme mentioned in this section does not create the parse table to parse a sentence. Hence it would be difficult to check the errors in a given sentence if it is not a string of the grammar. But in our approach, the errors can be detected by by means of the empty entry in the parse table.

# 3. A PARSING SCHEME FOR BANGLA GRAMMAR RECOGNITION

A predictive parser is an efficient way of implementing recursive decent parsing by handling the stack of activation record. The predictive parser has an input , a stack, a parse table and output. The input contains the string to be parsed or checked, followed by a $, the right end marker. The stack contains a sequence of grammar symbols, the parse table is a two dimensional array M[A,$n$] where A is nonterminal and n is a terminal or $ sign.

Table 1: Tag Set Description for Bangla grammar

| Tag Name (Symbol) | Examples |
|---|---|
| Noun (noun) | রহিম,বিদ্যালয়,ছেলে ,বই , ভাত,টাকা ,কানে,সিংহ ,ঘাস । |
| pronoun( pronoun) | আমি ,আমরা, তুমি ,সে , তারা |
| Adjective (adjective) | ভাল , বলবান, গুনবান ,একটু , দশ ,অনেক ,শখ,সর্বদা ,দ্রুত ,দীর্ঘতম । |
| verb(verb) | খাই ,খেলে ,করবে ,পড়া ,চলে ,যাই । |
| Conjunction(conjunction) | এবং ,ও, চেয়ে । |
| Negative Description(neg) | না ,নয় । |
| Modifier(modifier) | এ , একটি ,একদিন , এই । |

## A. Storing words in XML

The Extensible Markup Language (XML) is now used almost everywhere for its simplicity and ease of use. It is a good format to store any kind of data. The tasks behind using XML are always the same: reading data from XML and writing data into it. The general format of XML tag is  <tag_name>value</tag_name>.  A dictionary is a very basic NLP tool used to get the meaning, parts of speech, and usage of a word and can also be used as a spell-checker to detect errors in a sentence and correct them by providing a set of correct alternatives which includes the intended word. To design the dictionary in XML, we consider *Bangla word* as tag_name and *value* as its corresponding POS. An example of XML file is shown in Figure 1.

## B. Developing the tag set for Bangla grammar using XML

Each sentence is composed of one or more phrases. So if we can identify the syntactic constituents of sentences, it will be easier for us to obtain the structural representation of the sentence [8,9] . We tagged Bangla words with their respective parts-of-speech (POS), modifier, pattern, number [3] and stored in XML file. Table 1 shows the tag set used in XML file for storing Bangla words. Each sentence is partitioned into its constituent. A constituent expresses the complete meaning for a specific context.  The constituents of the sentences found out are shown in the table 2.

## C. Bangla Grammar Design

Once constituents have been identified, the productions for Context Free Grammar (CFG) are developed for Bangla sentence structures. Figure 2 shows the proposed grammar. As Bangla grammar has different forms, the same production term can be used only by reorganizing the in the grammar. For example, following three forms can be applied by reorganizing the production terms.

১.   আমি ভাত খাই।

২.   আমি খাই ভাত ।

৩.   ভাত আমি খাই ।

Table 2: Constituents for developing the grammar

| Constituents (Symbol) | Productions | Examples |
|---|---|---|
| Noun Phrase (NP) | NP -> noun<br>NP -> pronoun<br>NP ->modifier noun etc. | রহিম ,<br>আমি ,<br>এ পথে। |
| Verb Phrase(VP) | VP -> noun verb<br>VP -> noun verb verb<br>VP ->  noun verb ptrn etc. | বই পড়ছে,<br>পথ্য সেবন করে,<br>খাওয়া হল না। |
| Adjective Phrase(AP) | AP -> adjective<br>AP -> adjective noun etc . | দশ, ভাল ছেলে। |

*Left Factoring*: The grammar shown in Figure 2 is ambiguous. The parser generated from this kind of grammar is not efficient as it requires backtracking. To remove the ambiguity from the grammar we have used the idea of left factoring and reconstruct the grammar productions.

Left factoring is a grammar transformation useful for producing a grammar suitable for predictive parsing. The basic idea is that when it is not clear which of the productions are to use to expand a non terminal then it can defer to take decision until we get an input to expand it. In general, if we have productions of form

$A \rightarrow \alpha\beta_1 \mid \alpha\beta_2$

We left factored productions by getting the input α and break it as follows

$A \rightarrow \alpha A'$

$A' \rightarrow \beta_1 | \beta_2$

The above grammar is correct and is free from conflicts. After left factoring, the reconstructed grammar productions of Figure 2 are shown in Figure 3.

## D. Parser Design

A parser for a grammar G is a program that takes a string as input and produces a parse tree as output if the string is a sentence of G or produces an error message indicating that the sentence is not according to the grammar G. The idea of predictive parser design is well understood in a compiler design[11]. To construct a predictive parser for grammar G two functions namely FIRST() and FOLLOW() are important. These functions allow the entries of a predictive parse table for G. Once the parse table has been constructed we can verify any string whether it satisfy the grammar G or not. The FIRST() and FOLLOW() determines the entries in the parse table. Any other entries in the parse table are error entries.

*Computing FIRST of variable* : FIRST(α) be the set of terminals that begin the strings derived from α. If α → ε then ε is also included in FIRST(α). According the rules of computing FIRST(α) [6], the values for FIRST() are computed for the grammar G and are shown in Figure 4. The rules for computing FIRST of a terminal α is given in Appendix.

*Computing FOLLOW of a Non terminal*: FOLLOW(A) of a non terminal *A* is the set of terminals *a* that can appear immediately to the right of A. If A is the right most symbol in the sentential form then $ is added to FOLLW(A). According the rules of computing FOLLOW[6], the values are computed for the grammar G (see Figure 3) and are shown in Figure 5. The rules for computing FOLLOW of a non terminal *A* is given in Appendix.

---

*S->NP VP*

*NP->noun conjunction noun | noun ip| noun pronoun conjunction pronoun| noun pronoun ip| noun pronoun noun| noun pronoun adjective| noun pronoun pronoun| noun pronoun tp| noun adjective| noun noun conjunction pronoun| noun noun aw| noun noun| noun| noun pronoun| pronoun conjunction pronoun| pronoun ip| pronoun noun conjunction noun| pronoun noun ip| pronoun noun adjective| pronoun noun conjunction pronoun| pronoun noun aw| pronoun noun| pronoun adjective| pronoun pronoun conjunction pronoun| pronoun pronoun aw| pronoun pronoun| pronoun tp| pronoun| modifier noun| modifier adjective ptrn| modifier adjective| modifier pronoun| modifier conjunction adjective| modifier| adjective noun| adjective pronoun| adjective conjunction adjective| adjective ptrn| adjective| ip| tp| xp ip| xp pronoun conjunction pronoun| xp pronoun ip| xp pronoun noun| xp pronoun adjective| xp pronoun tp| xp adjective| xp noun conjunction pronoun| xp noun aw| xp noun| xp tp| xp aw| tp pronoun| tp adjective| tp ip| tp pronoun conjunction pronoun| tp pronoun noun| tp pronoun adjective| tp pronoun tp| tp noun conjunction pronoun| tp noun aw| tp aw.*

*AP->adjective noun| adjective pronoun| adjective ptrn*

*VP->noun verb| noun verb verb ptrn| noun verb verb adjective verb| noun verb verb adjective noun| noun verb verb adjective pronoun| noun verb verb adjective| noun verb verb noun adjective verb| noun verb verb noun ptrn| noun verb verb noun aw| noun verb verb noun adjective verb| noun verb verb noun verb ptrn| noun verb verb noun verb aw| noun verb verb noun verb adjective| noun verb verb noun ptrn| nounverb verb noun pronoun| noun verb verb noun| pronoun adjective verb| pronoun verb verb ptrn| pronoun verb verb adjective verb| pronoun verb verb adjective noun| pronoun verb verb adjective pronoun| pronoun verb verb ptrn| pronoun verb aw| pronoun verb adjective| pronoun verb pronoun| pronoun verb noun| pronoun ptrn | verb verb verb adjective| verb verb verb adjective verb| verb verb verb ptrn| verb verb verb pronoun| verb verb ptrn| verb verb adjective verb| verb verb adjective noun| verb verb adjective pronoun| verb verb noun adjective verb| verb verb noun verb ptrn| verb verb noun verb aw| verb verb noun verb adjective| verb verb noun ptrn| verb verb noun verb pronoun| verb adjective| verb adjective verb ptrn| verb adjective verb adjective verb| verb adjective verb noun| verb adjective verb noun verb | adjective noun verb adjective verb| adjective noun verb ptrn| adjective noun verb pronoun| adjective noun adjective ptrn| adjective noun adjective aw| adjective noun adjective | adjective pronoun| adjective ptrn | conjunction*

---

Figure 2: The proposed Bangla grammar G

*Parse Table Construction:* Let M[m, n] be a matrix where m is the number of non terminals in grammar G and n is the number of distinct input symbols that may occur in a sentence of grammar G. Table 3 shows the resulting parse table for the grammar G constructed by applying the following rules

1. for each production of the form A → α of the grammar G
   for each terminal *a* in first(α), add A → α to M[A,a].

2. If ε is in first(α) then
   for each terminal in FOLLOW(A) add A → α to M[A,a]

All other undefined entries of the parsing table are error entries.

```
S->NP VP

NP->noun NP1|pronoun NP2|modifier AP1|conjunction NP1| AP | NP2 |xp
NP1 |tp NP1

NP1->conjunction noun |ip | pronoun NP2 | adjective | noun NP3 | tp | aw |ε

NP2->conjunction pronoun | ip | noun NP1 | adjective | pronoun NP3 | tp | ε

NP3->conjunction pronoun| aw | ε

AP->adjective AP1

AP1->noun | adjective AP2 | pronoun | conjunction AP |ε

AP2->ptrn | ε

VP->noun VP1| pronoun VP4 | verb VP2 | AP VP3 | conjunction

VP1->verb VP2 | adjective VP3 | pronoun noun |noun pronoun

VP2->verb VP3 | ptrn | aw | AP VP3 |ε

VP3->verb VP4 | ptrn | adjective VP5 | noun VP4 |ε

VP4->AP verb | verb VP2| ptrn | pronoun

VP5->verb| noun VP4| pronoun AP1
```

Figure 3: Grammar after left factoring

```
FIRST(S) ={noun, pronoun, modifier, adjective, conjunction, xp, tp}
FIRST(NP) ={noun, pronoun, modifier, adjective, conjunction, xp, tp}
FIRST(NP1) ={conjunction,  ip, pronoun, adjective, noun, tp, aw, ε }
FIRST(NP2) = { conjunction,  ip, pronoun, adjective, noun, tp,  ε }
FIRST(NP3) = {conjunction, aw, ε}
FIRST(AP) ={adjective}
FIRST(AP1) = {noun, adjective, pronoun, conjunction, ε }
FIRST(AP2) = {ptrn, ε }
FIRST(VP) = {noun, pronoun, verb, adjective, conjunction}
FIRST(VP1) = {verb, adjective, pronoun, noun}
FIRST(VP2) = {verb, ptrn, aw, adjective, ε }
FIRST(VP3) = {noun, verb, adjective, ptrn, ε }
FIRST(VP4)={verb, ptrn, pronoun, adjective, ε }
FIRST(VP5)={verb, noun, pronoun}
```

Figure 4: Computing FIRST for the grammar G

*Parse Tree Generation*: A parse tree for a grammar G is a tree where the root is the start symbol for G, the interior nodes are the non terminals of G and the leaf nodes are the terminal symbols of G. The children of a node T (from left to right) correspond to the symbols on the right hand side of some production for T in G. Every terminal string generated by a grammar has a corresponding parse tree and every valid parse tree represents a string generated by the grammar. We store the parse table M using a two-dimensional array. To read an element from a two-dimensional array, we must identify the subscript of the corresponding *row* and then identify the subscript of the corresponding *column*. For example, the production "*NP→ modifier noun*" is in row 2, column 3, (see table 3) so it is identified as M[2][3].

```
FOLLOW(S) = { noun, adjective, pronoun, modifier, ip, xp, tp, conjunction }
FOLLOW(NP) ={noun, verb, adjective, pronoun}
FOLLOW(NP1) = {noun, verb, adjective, pronoun}
FOLLOW(NP2) = {noun, verb, adjective, pronoun}
FOLLOW(NP3) = {noun, verb, adjective, pronoun}
FOLLOW(AP) = {noun, verb, adjective, pronoun, ptrn, modifier, ip, xp, tp, conjunction}
FOLLOW(AP1) = {noun, verb, adjective, pronoun, ptrn, modifier, ip, xp, tp, conjunction}
FOLLOW(AP2) = {noun, verb, adjective, pronoun, ptrn, modifier, ip, xp, tp, conjunction}
FOLLOW(VP) ={ noun, adjective, pronoun, modifier, ip, xp, tp, conjunction, $}
FOLLOW(VP1) ={ noun, adjective, pronoun, modifier, ip, xp, tp, conjunction, $}
FOLLOW(VP2) ={ noun, adjective, pronoun, modifier, ip, xp, tp, conjunction, $}
FOLLOW(VP3) ={ noun, adjective, pronoun, modifier, ip, xp, tp, conjunction, $}
FOLLOW(VP4) ={ noun, adjective, pronoun, modifier, ip, xp, tp, conjunction, $}
FOLLOW(VP5) ={ noun, adjective, pronoun, modifier, ip, xp, tp, conjunction, $}
```

Figure 5: Computing FOLLOW for the grammar G

Let us explain the grammar for the sentence "একটি ছেলে বই পড়ছে". Using the XML data file we get the tags "*modifier noun noun verb*". Using the production $S \rightarrow NPVP$ of the grammar G (Figure 3) the sentence matches to *NP noun verb*; in the second iteration the *noun verb* part matches to *noun VP1*. The *VP1* in turn matches to *verb VP4* and *VP4* produces ε. Figure 6 shows the moves of our implementation using a stack.

| Stack | Input | Action |
|---|---|---|
| $ S | modifier  noun noun verb $ | |
| $ VP NP | modifier  noun noun verb $ | S->NP VP |
| $ VP noun modifier | modifier  noun noun verb $ | NP-> modifier noun |
| $ VP noun | noun noun verb $ | Poped |
| $ VP | noun verb $ | Poped |
| $ VP1 noun | noun verb $ | VP-> noun VP1 |
| $ VP1 | verb $ | Poped |
| $ VP2 verb | verb $ | VP1-> verb VP2 |
| $ VP2 | $ | Poped |
| $ | $ | VP2-> ε |
| $ | $ | Sentence is accepted |

Figure 6: Moves made by a Bangla parser on input "modifier noun noun verb" for correct sentence.

Initially, the parser is in a configuration with S$ in the input buffer and the start symbol S on top of the stack and then $. Figure 7 shows parsing algorithm which uses the parse table *M* of Table 3 to produce a Bangla parser for the input of a Bangla sentence. Figure 8 shows the snap shot of parse tree generated by the grammar G for the input "একটি ছেলে বই পড়ছে" of the form *modifier noun noun verb*.

*Managing Non Traditional Forms* : The structure of Bangla language may change in a sentence although the meaning does not change. For example, the sentence "আমি ভাত খাই " can also be written as আমি খাই ভাত or ভাত আমি খাই । The later two forms are also correct and have the same meaning. Hence it is sometimes difficult to detect such reorganized non traditional forms in a single grammar. The proposed grammar can detect the the non traditional forms. For example, the grammar G detects the forms "আমি খাই ভাত" and  "ভাত আমি খাই" are shown in Figure 9 and Figure 11 respectively.  Figure 10 and Figure 12 shows the corresponding parse tree for of Figure 9 and Figure 10.  Figure 13 shows the moves for an incorrect sentence "রহিম বিদ্যালয়ে" of the form *"noun noun"*.  The sentence is rejected because the there is no verb in the sentence and such types of productions are not preseneted in the grammar G.

1.  set Input Pointer(IP) to point to the first word of w;
2.  set X to the top stack word;
3.  **while** ( X!= $ ) **begin**  /* stack is not empty */
    **a.**  if  X is a terminal,  pop the stack and advance IP;
    **b.**  if X is a Nonterminal and  M[X,IP] has  the production  X → Y1Y2…Yk
          output the production X -> Y1Y2 -Yk;
                pop the stack;
                push Yk, Yk-1,. . . , Y1 onto the stack, with Y1 on top;
    **c.**  if X=$ ,Sentence is Accepted.
          **end**

Figure 7:  Bangla parsing algorithm

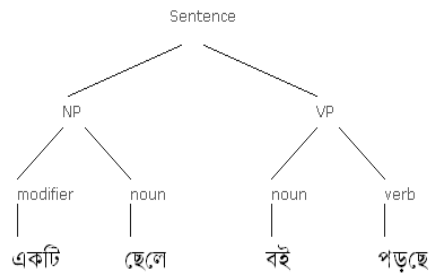

Figure 8: Parse tree for Bangla parser on input *"modifier noun noun verb"*

| Stack | Input | Action |
|---|---|---|
| $ S | pronoun verb noun $ | |
| $ VP NP | pronoun verb noun $ | S->NP VP |
| $ VP NP2 pronoun | pronoun verb noun $ | NP->pronoun NP2 |
| $ VP NP2 | verb noun $ | poped |
| $ VP | verb noun $ | NP2->zero |
| $ VP3 verb | verb noun $ | VP->verb VP3 |
| $ VP3 | noun $ | poped |
| $ noun | noun $ | VP3->noun |
| $ | $ | poped |
| $ | $ | Sentence is accepted |

Figure 9: Moves made by a Bangla parser on input "pronoun verb noun" for correct sentence "আমি খাই ভাত"

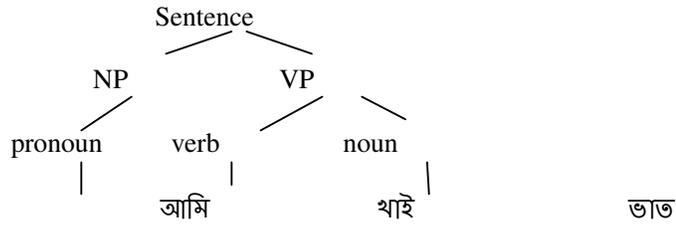

Figure 10: Parse tree for Bangla parser on input *"pronoun verb noun"*.

| Stack | Input | Action |
|-------|-------|--------|
| $ S | noun pronoun verb $ | |
| $ VP NP | noun pronoun verb $ | S->NP VP |
| $ VP NP1 noun | noun pronoun verb $ | NP->noun NP1 |
| $ VP NP1 | pronoun verb $ | poped |
| $ VP | pronoun verb $ | NP1->zero |
| $ verb AP pronoun | pronoun verb $ | VP->pronoun AP verb |
| $ verb AP | verb $ | poped |
| $ verb | verb $ | AP->zero |
| $ | $ | poped |
| $ | $ | Sentence is accepted |

Figure 11: Moves made by a Bangla parser on input "noun pronoun verb" for correct sentence "ভাত আমি খাই".

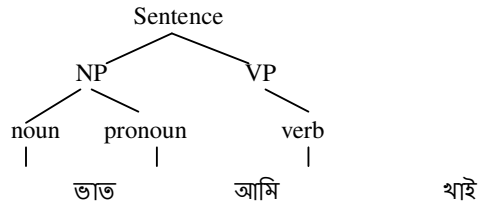

Figure 12: Parse tree for Bangla parser on input *"noun pronoun verb"*.

| Stack | Input | Action |
|-------|-------|--------|
| $ S | noun noun $ | |
| $ VP NP | noun noun $ | S->NP VP |
| $ VP NP1 noun | noun noun $ | NP->noun NP1 |
| $ VP NP1 | noun $ | Poped |
| $ VP NP3 noun | noun $ | NP1->noun NP3 |
| $ VP NP3 | $ | Poped |
| | | Sentence is rejected |

Figure 13: Moves made by a Bangla parser on input "noun noun" for incorrect sentence.

Table 3: Predictive Parsing Table for the developed Bangla grammar G

| | noun | pronoun | modifier | adjective | Verb | conjunction | ptrn | ip | aw | xp | tp | $ |
|---|---|---|---|---|---|---|---|---|---|---|---|---|
| S | NP VP | NP VP | NP VP | NP VP | | NP VP | | NP VP | | NP VP | NP VP | |
| NP | noun NP1 | pronoun NP2 | modifier noun | AP | | conjunction NP1 | | NP2 | aw | xp NP1 | tp NP1 | |
| NP1 | noun NP3 | pronoun NP2 | NP | adjective | ε | conjunction VP2 | | ip | | | tp | |
| NP2 | noun NP1 | pronoun NP3 | | adjective | | conjunction pronoun | | ip | | | tp | |
| NP3 | ε | ε | | ε | ε | conjunction AP | | | aw | | | |
| AP | | | | adjective AP1 | | | | | | | | |
| AP1 | noun | pronoun | ε | adjective AP2 | ε | conjunction AP | ε | ε | | ε | ε | ε |
| AP2 | ε | ε | ε | ε | ε | ε | ptrn | | ε | ε | ε | ε |
| VP | noun VP1 | pronoun VP4 | | AP VP3 | verb VP2 | conjunction | | | | | | |
| VP1 | noun pronoun | pronoun Noun | | adjective noun | verb VP2 | | | | | | | |
| VP2 | noun VP3 | ε | ε | AP VP3 | verb VP3 | ε | ptrn | ε | aw | ε | ε | ε |
| VP3 | noun VP4 | ε | ε | adjective VP5 | verb VP4 | ε | ptrn | ε | | ε | ε | ε |
| VP4 | | pronoun | | AP verb | verb VP2 | ε | ptrn | ε | | ε | ε | ε |
| VP5 | noun VP4 | pronoun | AP1 | | Verb | | | | | | | |

# 4.  EXPERIMENTAL RESULTS

In this Section we show some input sentences that is used for performance analysis. We have used three types of sentences namely simple and traditional form, some nontraditional form and paragraphs. The paragraphs are collected from newspaper. By nontraditional form we mean the same meaning of another sentence having structural similarity. For example, "আমি

ভাত খাই" and "খাই আমি ভাত" sentences are approximately similar but later on is rarely used. Table 4 shows the success rate of our proposed grammar.

  a)  Input sentences

তিনি কি ভাল না?,  আমি ভাত খাই। আমি কি ভাত খাই?, আমি কি ভাত খাই না?  "বাহ! পাখিটি তো খুব সুন্দর"। শিতে আমরা খুব কষ্ট পাই । তোমরা আগামীকাল এসো। আমি,তুমি ও সে বাড়ি  যাই।  আগামীকাল কি তুমি  আসবে?

**b) Input Paragraphs**

**i)** মাঝরাতে হঠাৎ করেই আপনার অসুখ করল। চিকিৎসক বা হাসপাতাল আশপাশে নেই। ভয় পাওয়ার কোন কারণ নেই। যদি আপনার হাতের কাছেই মোবাইল পান,তবে ডায়াল করুন হেল্প লাইনে। শীঘ্রই আপনি পেয়ে যাবেন দরকারি পরামর্শ।

**ii)** আমরাই একমাত্র জাতি যারা ভাষার জন্য যুদ্ধ করেছি। মায়ের ভাষা বাংলাকে ছিনিয়ে এনেছি। ভাষার এই যুদ্ধ এখানেই বন্ধ হয়নি। বাংলাকে আলাদা করে প্রতিষ্ঠিত করার জন্য শুরু হয়েছে প্রযুক্তির ব্যাবহার। সেই যুদ্ধে ঝাঁপিয়ে পড়েছে মেধাবী জনতা।

**iii)** একসময় বাংলা ভাষা যে হুমকির মুখে ছিল, তার মূলে ছিল উপনিবেশিক শাসন। যা ছিল সংখ্যাগুরুর ওপর চাপিয়ে দেওয়া সংখ্যালঘুর শাসন। তার বিরুদ্ধে লড়াই করেছিল বাংলা ভাষাভাষী বেশির ভাগ মানুষ। কিন্তু একটি ভাষার শক্তি এর ব্যাবহারকারীদের মোট শক্তির সমান। ফলে বাংলা ভাষার বিরুদ্ধকারীরা পরাজিত হয়েছে। যেহেতু তারা সংখ্যায় বেশি ছিল না।

Table 4: Success rate for different Bangla sentences

| Types of sentences | Total no. of sentences/ paragrph (I) | Correctly detected (D) | Acceptance Rate A=(D/I)*100% |
|---|---|---|---|
| Traditional | 120 | 100 | 83.33% |
| Nontraditional | 80 | 58 | 72.5% |
| Paragraph | 25 | 18 | 72% |

# 5. CONCLUSION

In this paper we describe a context free grammar for Bangla language and hence we develop a Bangla parser based on that grammar. Our approach is very much general to apply in Bangla Sentences and the method is well accepted for parsing a language of a grammar. We have presented a scheme which is effective for detecting a paragraph contains of almost all type of traditional and non traditional Bangla sentences. The structural representation that has been built can cover the maximum simple, complex and compound sentences. But there are some sentences composed of idioms and phrases are beyond the scope of this paper. Also mixed sentences are of out of the discussion. But further increasing and modifying the production rule it can be possible to remove the above limitations .We believe the proposed method can be applied to check most of the Bangla grammar to parse Bangla language.

## Appendix

### Rules for computing FIRST

- If X is a terminal symbol then FIRST(X)={X}
- If X is a nonterminal symbol and X $\rightarrow$ ε is a production rule then ε is in first(X).
- If X is a non-terminal symbol and X $\rightarrow$ $Y^1Y^2...Y^n$ is a production rule then first(X)=first(Y1).

### Rules for computing FOLLOW

- If S is the start symbol then $ is in FOLLOW(S)
- If A $\rightarrow$ αBβ is a production rule then everything in FIRST(β) is FOLLOW(B) except ε.
- If ( A $\rightarrow$ αB is a production rule ) or ( A $\rightarrow$ αBβ is a production rule and ε is in first(β) ) then everything in FOLLOW(A) is in FOLLOW(B).

## Authors

Prof. Dr. K. M. Azharul Hasan received his B.Sc. (Engg.) from Khulna University, Bangladesh in 1999 and M.E. from Asian Institute of Technology (AIT), Thailand in 2002 both in Computer Science. He received his Ph.D. from the Graduate School of Engineering, University of Fukui, Japan in 2006. His research interest lies in the areas of databases and software engineering, and his main research interests include Parallel and distributed databases, Parallel algorithms, NLP, Information retrieval, Data warehousing, MOLAP, Multidimensional databases, OOAD, Software metric and Software maintenance. He is with the Department of Computer Science and Engineering, Khulna University of Engineering and Technology (KUET), Bangladesh since 2001.

Al-Mahmud received his B.Sc. (Engg.) in Computer Science and Engineering from Khulna University of Engineering and Technology (KUET), Bangladesh in 2010. Currently he is working as a lecturer in the same department. His research interests lies in the area of Soft Computing, Bio-inspired Computing, Natural Language Processing (NLP), Graph Theory, Cryptography and Soft Error Tolerance. Currently he is servicing as a lecturer of Department of Computer Science and Engineering in Khulna University of Engineering and Technology (KUET), Bangladesh.

Amit Mondal received his B.Sc. (Engg.) in Computer Science and Engineering from Khulna University of Engineering and Technology (KUET), Bangladesh in 2010. His research interests lies in the area of Natural Language Processing (NLP. Currently he is servicing as a Software Engineer in Bangladesh Internet Press Limited, Bangladesh.

Amit Saha received his B.Sc. (Engg.) in Computer Science and Engineering from Khulna University of Engineering and Technology (KUET), Bangladesh in 2010. His research interests lies in the area of Natural Language Processing (NLP. Currently he is servicing as a Junior Software Engineer in Adharsoft Incorporation, Bangladesh.